\documentclass[letterpaper]{article} 
\usepackage{aaai24}  
\usepackage{times}  
\usepackage{helvet}  
\usepackage{courier}  
\usepackage[hyphens]{url}  
\usepackage{graphicx} 
\usepackage{natbib}  
\usepackage{caption} 
\usepackage{algorithm}
\usepackage{algorithmic}

\usepackage{amssymb}
\usepackage{amsmath}
\usepackage{graphicx}
\usepackage{subfigure}
\usepackage{threeparttable}
\usepackage{multirow}
\usepackage{booktabs}

\usepackage{amsmath}

\usepackage{makecell}

\usepackage{subfigure}
\usepackage{threeparttable}
\usepackage{multirow}
\usepackage{bm}
\usepackage{xcolor}

\definecolor{uncertain}{RGB}{255, 0, 0}

\usepackage{soul}
\definecolor{green}{RGB}{0, 100, 0}
\definecolor{red}{RGB}{139, 0, 0}
\definecolor{gold}{RGB}{255, 125, 0}
\definecolor{grey}{RGB}{200, 200, 200}
\definecolor{white}{RGB}{255, 255, 255}
\definecolor{orange}{RGB}{255, 228, 201}
\definecolor{blue}{RGB}{207, 226, 243}
\definecolor{darkblue}{RGB}{47,85,151}
\definecolor{darkred}{RGB}{192,0,0}

\usepackage{tabularx}

\usepackage{newfloat}
\usepackage{listings}
\DeclareCaptionStyle{ruled}{labelfont=normalfont,labelsep=colon,strut=off} 
\floatstyle{ruled}
\newfloat{listing}{tb}{lst}{}
\floatname{listing}{Listing}
\title{Can Large Language Models Serve as Rational Players in Game Theory?\\A Systematic Analysis}

\author {
    Caoyun Fan\thanks{These authors contributed equally. },
    Jindou Chen\footnotemark[1],
    Yaohui Jin\thanks{Corresponding author. },
    Hao He\footnotemark[2]
}
\affiliations {
    MoE Key Lab of Artificial Intelligence, AI Institute, Shanghai Jiao Tong University\\
    \texttt{\{fcy3649, goldenbean, jinyh, hehao\}@sjtu.edu.cn}
}

\begin{document}

\maketitle

\begin{abstract}

Game theory, as an analytical tool, is frequently utilized to analyze human behavior in social science research. With the high alignment between the behavior of Large Language Models (LLMs) and humans, a promising research direction is to employ LLMs as substitutes for humans in game experiments, enabling social science research. However, despite numerous empirical researches on the combination of LLMs and game theory, the capability boundaries of LLMs in game theory remain unclear. In this research, we endeavor to systematically analyze LLMs in the context of game theory. Specifically, rationality, as the fundamental principle of game theory, serves as the metric for evaluating players' behavior --- building a clear desire, refining belief about uncertainty, and taking optimal actions. Accordingly, we select three classical games (dictator game, Rock-Paper-Scissors, and ring-network game) to analyze to what extent LLMs can achieve rationality in these three aspects. The experimental results indicate that even the current state-of-the-art LLM (GPT-4) exhibits substantial disparities compared to humans in game theory. For instance, LLMs struggle to build desires based on uncommon preferences, fail to refine belief from many simple patterns, and may overlook or modify refined belief when taking actions. Therefore, we consider that introducing LLMs into game experiments in the field of social science should be approached with greater caution. 

\end{abstract}

\section{Introduction}
\label{s1}

Game theory \cite{Roughgarden2010AlgorithmicGT,Dufwenberg2011GameT} is a mathematical theory for evaluating human behavior. Due to its highly abstract representation of real-life situations \cite{Osborne1995ACI}, it becomes a standard analytical tool \cite{charness2002understanding,cachon2006game} in the field of social science (e.g., economics, psychology, sociology, etc.). With the rapid development of Large Language Models (LLMs) \cite{DBLP:conf/nips/Ouyang0JAWMZASR22,DBLP:journals/corr/abs-2303-08774}, a significant advancement is the high alignment between the behavior of LLMs and humans \cite{DBLP:journals/corr/abs-2204-05862,DBLP:conf/nips/Ouyang0JAWMZASR22,ICASSP2024}. As a result, many researchers consider LLMs as human-like research subjects \cite{Dillion2023CanAL} and analyze LLMs’ professional competence in social science through game experiments \cite{chen2023emergence,DBLP:journals/corr/abs-2305-16867,johnson2023evidence}. However, despite the strong motivation for the combination of LLMs and game theory \cite{Horton2023LargeLM,Guo2023GPTAI}, the preliminary researches mainly treat LLMs and game theory empirically as analytical tools in social science \cite{Aher2022UsingLL,park2022social,DBLP:journals/corr/abs-2305-16867,bybee2023surveying}, without systematically analyzing LLMs in the context of game theory. As a result, many fundamental aspects of LLMs in game theory remain unclear. For example, what research subjects cannot LLMs play? What types of games are LLMs not good at playing? What kind of game processes are LLMs more suitable for? And so on. 

\begin{figure}[t]
    \centering
    \includegraphics[width=215pt]{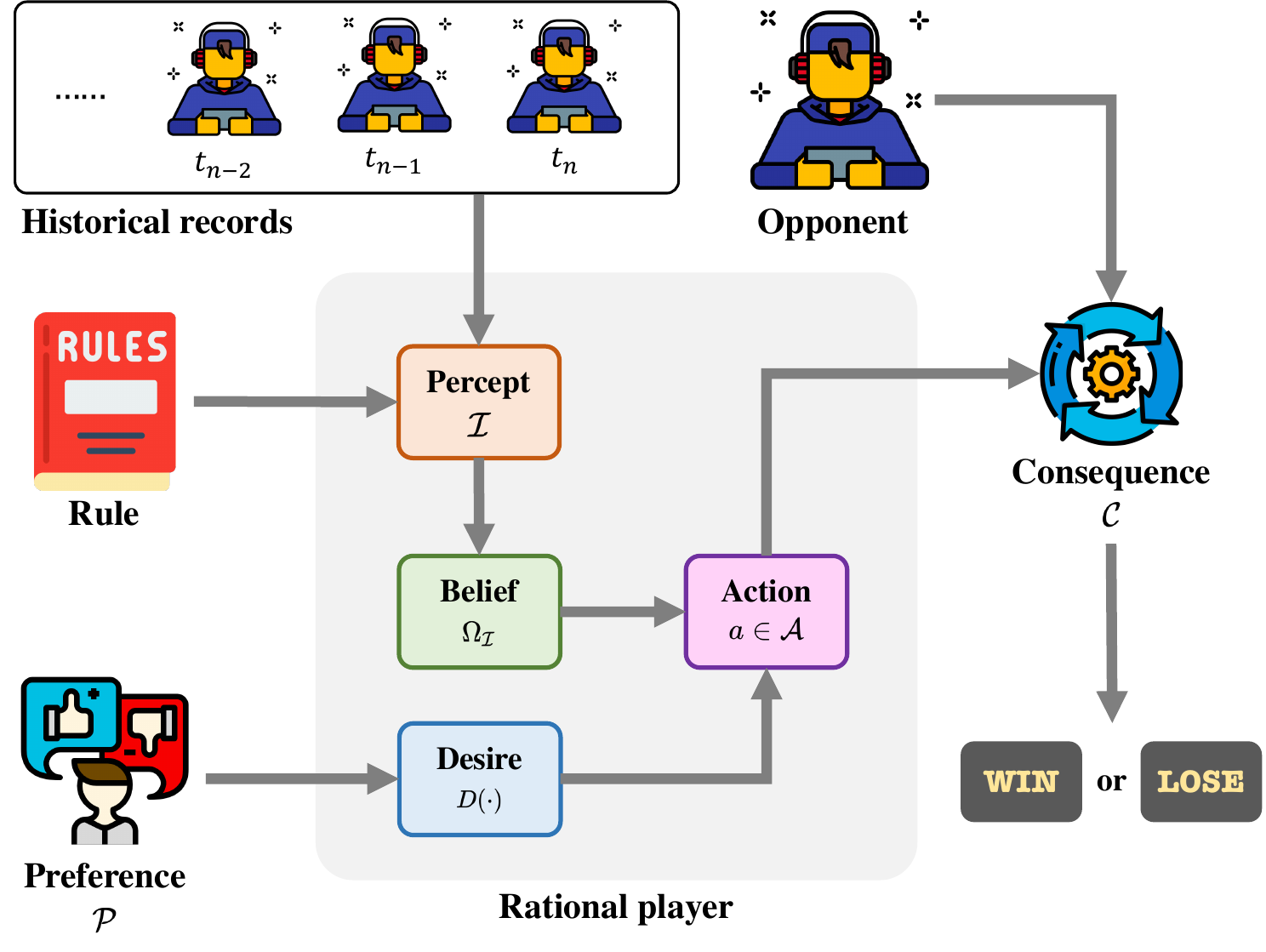}
    \caption{Overview of a player's behavior in game theory. }
    \label{f1-1}
\end{figure}

We consider it necessary to systematically analyze LLMs in the context of game theory, because such analysis can clarify the capability boundaries of LLMs and provide further guidance for the widespread use of LLMs in social science research. Essentially, the role of game theory is to evaluate the behavior of the research subjects (players) \cite{Roughgarden2010AlgorithmicGT}, as shown in Fig. \ref{f1-1}, a player needs to take an action $a \in \mathcal{A}$ based on preference $\mathcal{P}$ and perceived game information $\mathcal{I}$ (e.g., game rules and historical records) in order to win the game. And rationality, as the fundamental principle of game theory, is the metric for evaluating players' behavior \cite{Roughgarden2010AlgorithmicGT,Dufwenberg2011GameT}. A rational player is considered to possess three characteristics \cite{Zagare1984GameTC,Osborne1995ACI} as: 
\begin{itemize}
    \item \emph{build a clear desire for the game. }
    \item \emph{refine belief about uncertainty in the game. }
    \item \emph{take optimal actions based on desire and belief. }
\end{itemize}

\noindent Specifically, desire $D(\cdot)$ represents a player's (concrete) opinion of each consequence within a game, determined by a player's (abstract) preference $\mathcal{P}$. Belief $\Omega_\mathcal{I}$ is refined from the game information $\mathcal{I}$, and represents a player's subjective judgment of uncertainty (e.g., opponent's action). Taking the optimal action $a \in \mathcal{A}$ requires a player to reason by combining desire $D(\cdot)$ and belief $\Omega_\mathcal{I}$ in the game process. More details can be found in Section \ref{s3}. 

In this research, we consider the three characteristics of a rational player as a reasonable perspective for systematically analyzing LLMs in the context of game theory. Accordingly, we select three classical games (dictator game, Rock-Paper-Scissors, and ring-network game) for these three characteristics, respectively. With the dictator game, we find that LLMs have the basic ability to build a clear desire. However, when assigned uncommon preferences, LLMs often suffer from decreased mathematical ability and inability to understand preferences. With Rock-Paper-Scissors, we observe that LLMs cannot refine belief from many simple patterns, which makes us pessimistic about LLMs playing games that require refining complex beliefs. Nonetheless, GPT-4 exhibits astonishingly human-like performance in certain patterns, able to become increasingly confident of refined belief as the game information increases. With the ring-network game, we consider that LLMs cannot autonomously follow the player's behavior in Fig. \ref{f1-1}. Explicitly decomposing the behavior in the game process can improve the ability of LLMs to take optimal actions, but the phenomenon of overlooking / modifying refined belief remains unavoidable in LLMs. 

In summary, our research systematically explores the capability boundaries of LLMs in the context of game theory from three perspectives. We believe that our research can pave the way for the smooth introduction of LLMs in the field of social science. 

\section{Related Work}
\label{s2}

\subsection{LLMs in Social Science}
\label{s2-1}

A significant advantage of LLMs is the high alignment with human behavior \cite{DBLP:journals/corr/abs-2204-05862,DBLP:conf/nips/Ouyang0JAWMZASR22}. Therefore, from the perspective of cost and efficiency, many social science researches began to employ LLMs to replace humans as research subjects \cite{Aher2022UsingLL,argyle2023out,bybee2023surveying,park2022social}. For example, in order to explore fairness and framing effects in sociology, LLMs were introduced into the classic game experiments \cite{Horton2023LargeLM}, which demonstrated the potential of LLMs to deal with social issues; in the research of consumer behavior \cite {brand2023using}, the behavior of LLMs was consistent with economic theory in many respects (i.e. downward-sloping demand curves, diminishing marginal utility of income, and state dependence); in finance research \cite{chen2023emergence}, LLMs' decisions in budgetary allocation scenarios received higher rationality scores compared to humans; and in psychology experiments \cite{Dillion2023CanAL}, the behavior of LLMs was highly consistent with the mainstream values of society. 

While these researches demonstrate the rationality of LLMs replacing human research subjects in certain social science domains (in certain experiments), there is still a lack of systematic analysis of the capability boundaries of LLMs in social science. 





\subsection{Game Theory}
\label{s2-2}

Game theory, as a mathematical theory, provides a framework for analyzing and predicting the behavior of rational players under conditions of uncertainty \cite{Roughgarden2010AlgorithmicGT,Dufwenberg2011GameT}. Game theory was originally developed in economics \cite{ichiishi2014game}, and a wide range of economic behaviors, such as market competition, auction mechanism, and pricing strategies, were modeled as game experiments \cite{samuelson2016game}. With the rapid cross-fertilization of scientific theories \cite{Shubik1982GameTI}, game theory was also applied to politics, sociology, psychology, and other fields of social science \cite{larson2021networks}. 

The research on the performance of LLMs in game theory has the following advantages: strong operability, the experimental design of game theory is often relatively simple; strong analyzability, game theory has comprehensive theoretical support for the experimental results; strong generalization, game theory is a high-level abstraction of many phenomena in the field of social science. 


\section{Preliminaries of Game Theory}
\label{s3}

The core of game theory \cite{Roughgarden2010AlgorithmicGT} is to guide players to take optimal actions under conditions of uncertainty\footnote{We assume that uncertainty arises only from the opponent's action. All games in this research satisfy this assumption. }. Generally, a game is modeled in five parts: 
\begin{itemize}
    \item Game information $\mathcal{I}$, e.g., game rules, historical records. 
    \item A set $\mathcal{A}$ of actions from that players can take. 
    \item A set $\mathcal{C}$ of possible consequences of action. 
    \item A consequence function $g: \mathcal{A} \to \mathcal{C}$ that associates a consequence with each actions. 
    \item A desire function $D_c: \mathcal{C} \to \mathbb{R}$, which is determined by the player's preference $\mathcal{P}$. For any $c_1, c_2 \in \mathcal{C}$, the player prefers $c_1$ if and only if $D_c(c_1) > D_c(c_2)$. 
\end{itemize}

To eliminate uncertainty in the game process, almost all game researches employ the belief theory \cite{Morgenstern1945TheoryOG,Lindley1955TheFO}. That is, a rational player will estimate a (subjective) probability distribution for any uncertainty based on $\mathcal{I}$, and this is referred to as the player's belief \cite{Osborne1995ACI}. Specifically, the player is assumed to have a belief $\Omega_\mathcal{I}$, a belief's probability distribution $p(\Omega_\mathcal{I})$, a consequence function $g: \mathcal{A} \times \Omega_\mathcal{I} \to C$. Then, the player attempts to find the optimal strategy $\pi^*(a | \mathcal{I})$ by maximizing the expected desire with the consideration of $\Omega_\mathcal{I}$ as: 
\begin{equation}
    \pi^*(a | \mathcal{I}) = \operatorname*{argmax}_{\ a \in \mathcal{A}} \mathbb{E}_{\omega \sim p(\Omega_\mathcal{I})}[D(a, \omega)], 
    \label{e3-1}
\end{equation}

\noindent where $D(\cdot)$ is a simplification of $D_c \circ g(\cdot)$. 

In fact, Eq. \ref{e3-1} explicitly expresses three characteristics of a rational player: having a clear desire corresponds to building the desire function $D(\cdot)$; refining belief about uncertainty corresponds to sampling in the belief's probability distribution $\omega \sim p(\Omega_\mathcal{I})$; taking optimal actions corresponds to choosing the action that maximizes desire $\underset{{\ a \in \mathcal{A}}}{\operatorname*{argmax}}D(a)$. 


\section{LLMs in Game Theory}
\label{s4}

In this section, we endeavor to conduct a systematic analysis of LLMs in the context of game theory. Specifically, we evaluate to what extent LLMs can achieve three characteristics of a rational player through three classic games (dictator game, Rock-Paper-Scissors, and ring-network game). The LLMs we analyze are openAI's \texttt{text-davinci-003} (GPT-3), \texttt{gpt-3.5-turbo} (GPT-3.5), \texttt{gpt-4} (GPT-4), the current state-of-the-art LLMs\footnote{\url{https://platform.openai.com/}}. All prompts used in the three games, as well as some examples of LLMs performance, can be found in Appendix. 

\subsection{Can LLMs Build A Clear Desire?}
\label{s4-1}

The premise of game theory is that each player has an abstract preference $\mathcal{P}$ for the consequence set $\mathcal{C}$. A rational player should build a concrete desire function $D(\cdot)$ based on preference $\mathcal{P}$ to measure the desire for each consequence $c \in \mathcal{C}$. In sociological research \cite{Burns2021SociologicalGT}, game experiments are frequently designed to explore the phenomenon where players with different preferences (cooperative or competitive) may have entirely different desires for the same consequence (win-win). 

For humans, preference and desire seem to be coexistent, while for LLMs, preference is assigned through a textual prompt. Therefore, we need to analyze whether LLMs can build reasonable desires from textual prompts. 


\subsubsection{Game: Dictator Game}

The dictator game \cite{charness2002understanding} is a classic game experiment in sociology \cite{Guala2010ParadigmaticET}, which is used to analyze players' personal preferences $\mathcal{P}$. In this game, there are two players: the dictator and the recipient. Given two allocation options, the dictator needs to take action, choosing one of two allocation options, while the recipient must accept the allocation option chosen by the dictator. Here, the dictator's choice is considered to reflect the personal preference \cite{Camerer1995AnomaliesUD,Leder2018}. For example, given two allocation options as: 
\begin{itemize}
    \item Option X: \emph{The dictator gets \$300, the recipient gets \$300. }
    \item Option Y: \emph{The dictator gets \$500, the recipient gets \$100. }
\end{itemize}

\noindent A dictator who prefers equality is more likely to choose Option X, while a dictator who prefers self-interest is more likely to choose Option Y. 

We choose the dictator game to analyze LLMs' desire for two reasons. First, the desires of this game are diverse. Unlike most games with a fixed preference (e.g., to maximize one's own interest), this game allows players to have diverse preferences, which results in diverse desire functions and different choices. Second, since the recipient's action is known (to accept), there is no uncertainty in this game, i.e., the belief $\Omega_\mathcal{I}$ is fixed to $\omega_\mathcal{I}$. This makes LLMs immune to potential interference from the biased belief. Therefore, the optimal strategy of the dictator game is expressed as: 
\begin{equation}
\pi^*(a | \mathcal{I}) = \operatorname*{argmax}_{a \in  \{X, Y\}} \{ D(X, \omega_\mathcal{I}), D(Y, \omega_\mathcal{I}) \}, 
\end{equation}

\noindent where $X$ and $Y$ refer to the dictator choosing option X and option Y, respectively. Thus, by providing multiple allocation options, we can analyze whether the desires built by LLMs match the corresponding preferences. 



\begin{table}[t]
    \centering
    \begin{threeparttable}
    \begin{tabular}{llcccc}
    \toprule
    \multirow{2}*{\bf LLM} & \multirow{2}*{\bf Pref.} & \multicolumn{4}{c}{\bf Option} \cr
    \cmidrule(lr){3-6}
      & & EQ & CI & SI & AL \cr
    \midrule
        \multirow{4}*{\bf GPT-3} & EQ & - & 1.0 & 1.0 & 1.0 \cr
         & CI & 0.4 & - & 0.3 & 0.5 \cr
         & SI & 1.0 & 1.0 & - & 1.0 \cr
         & AL & 0.0 & 0.0 & 0.1 & - \cr
    \midrule
        \multirow{4}*{\bf GPT-3.5} & EQ & - & 1.0 & 1.0 & 1.0 \cr
         & CI & 1.0 & - & 0.9 & 1.0 \cr
         & SI & 1.0 & 1.0 & - & 1.0 \cr
         & AL & 1.0 & 0.6 & 0.8 & - \cr
    \midrule
        \multirow{4}*{\bf GPT-4} & EQ & - & 1.0 & 1.0 & 1.0 \cr
         & CI & 1.0 & - & 1.0 & 0.9 \cr
         & SI & 1.0 & 1.0 & - & 1.0 \cr
         & AL & 1.0 & 1.0 & 1.0 & - \cr
    \bottomrule
    \end{tabular}
    \end{threeparttable}
    \caption{Accuracy of LLMs in the dictator game, where \textbf{Pref.} is an abbreviation for Preference. }
    \label{tab4-1-1}
\end{table}

\subsubsection{Setup}

Following \cite{Grech2018RationalAO}, we set four preferences for LLMs, to analyze different desires as: 
\begin{itemize}
    \item Equality (EQ): \emph{You have a stronger preference for fairness between players and hate inequality. } 
    \item Common-Interest (CI): \emph{You have a stronger preference for common interest and maximize the joint income. }  
    \item Self-Interest (SI): \emph{You have a stronger preference for your own interest and maximize your own income. } 
    \item Altruism (AL): \emph{You have a stronger preference for another player's interest and maximize another player's income. } 
\end{itemize}

\noindent Compared to the original setting \cite{charness2002understanding}, we adjust the allocation options corresponding to each preference to be closer and introduce an additional preference AL, thereby increasing the challenge of the game. Specifically, we set up allocation options for EQ, CI, SI, and AL as follows: (\$300, \$300), (\$400, \$300), (\$100, \$500), and (\$500, \$100), respectively. In each option, the first number represents the dictator's income, and the second number represents the recipient's income. It is worth noting that in game theory \cite{Osborne1995ACI}, SI and EQ are the most common preferences, followed by CI, while AL hardly ever occurs. 

In our experiments, we assign LLMs a specific preference (e.g., EQ) through a textual prompt, and then verify whether LLMs can make preference-consistent choices under different combinations of allocation options (i.e., EQ-CI, EQ-SI, and EQ-AL). Therefore, for each preference, each LLM is required to play three different dictator games. Each experiment is repeated 10 times and we report the accuracy. The temperature of LLMs is set to 0.7. 


\begin{figure}[t]
    \centering
    \includegraphics[width=239pt]{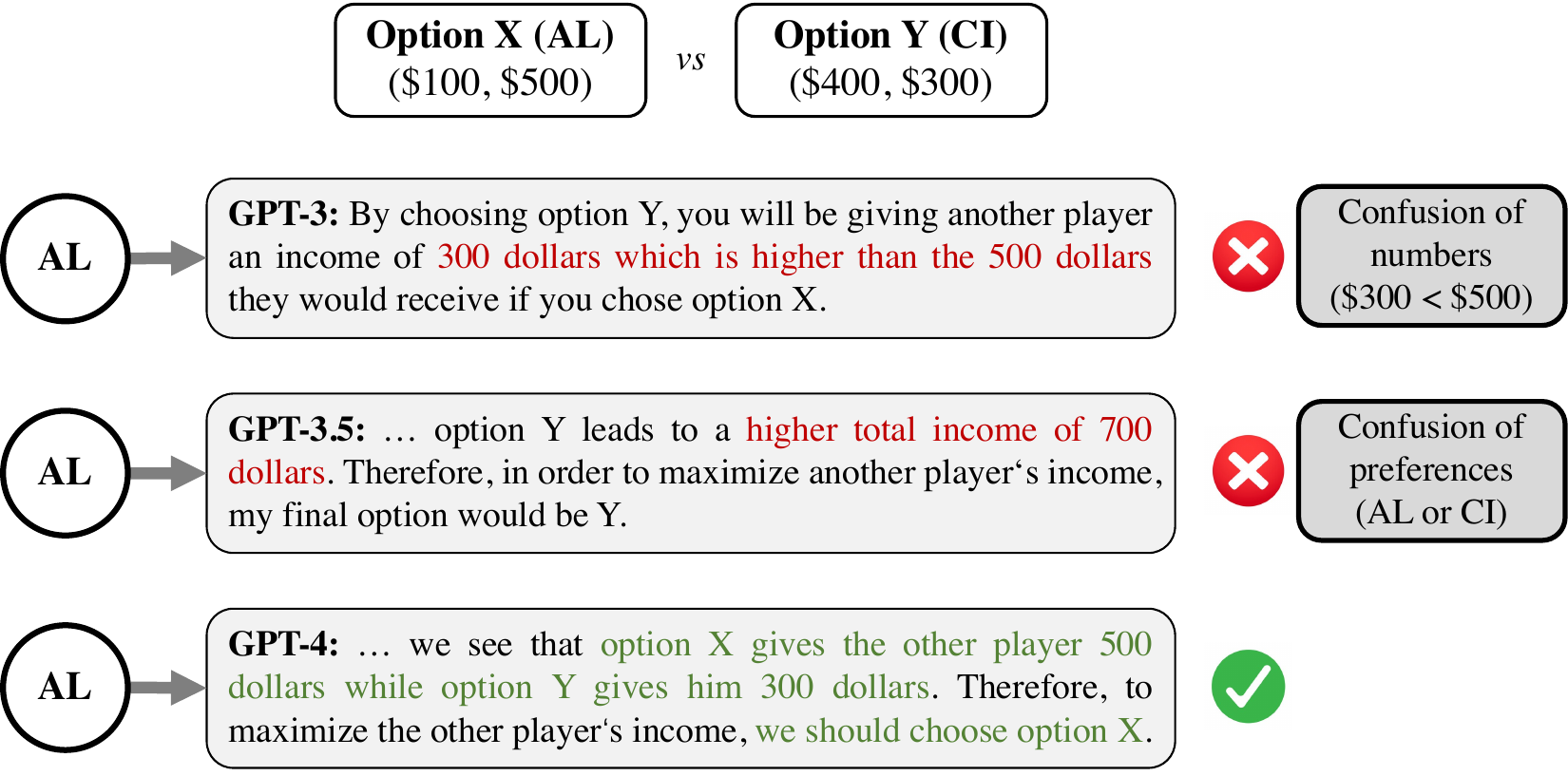}
    \caption{A case of the dictator game. All LLMs are assigned the preference AL, and the allocation options are AL-CI. }
    \label{f4-1-1}
\end{figure}

\begin{figure*}[t]
    \centering
    \subfigure[constant]{
    \includegraphics[width=0.323\columnwidth]{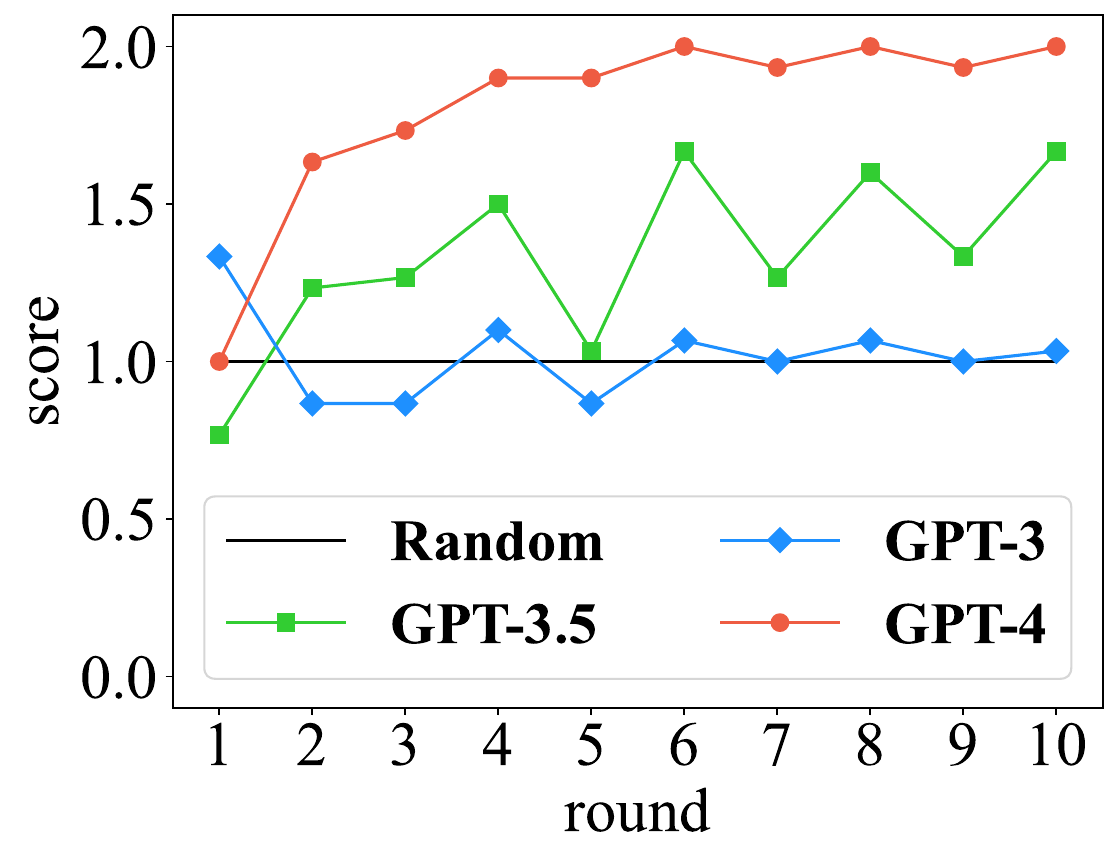}
    \label{f4-2-1a}
    }
    \subfigure[loop-2]{
    \includegraphics[width=0.323\columnwidth]{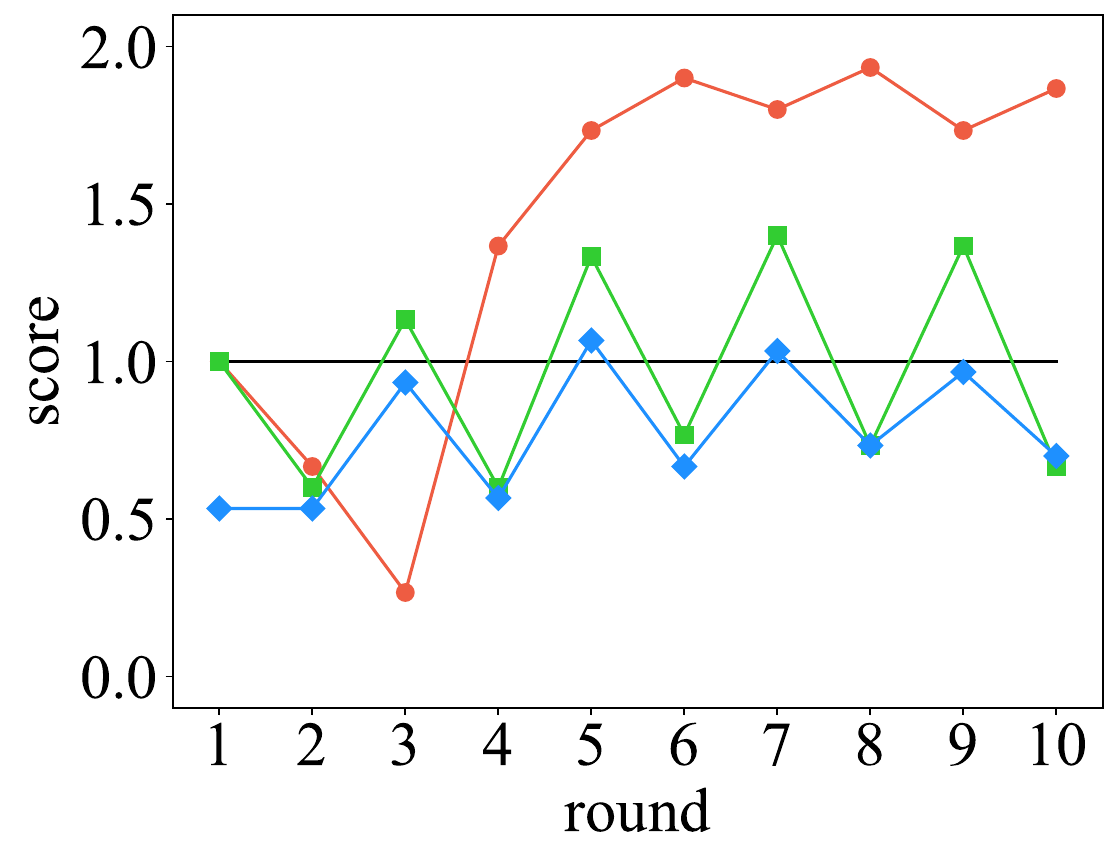} 
    \label{f4-2-1b}
    }
    \subfigure[loop-3]{
    \includegraphics[width=0.323\columnwidth]{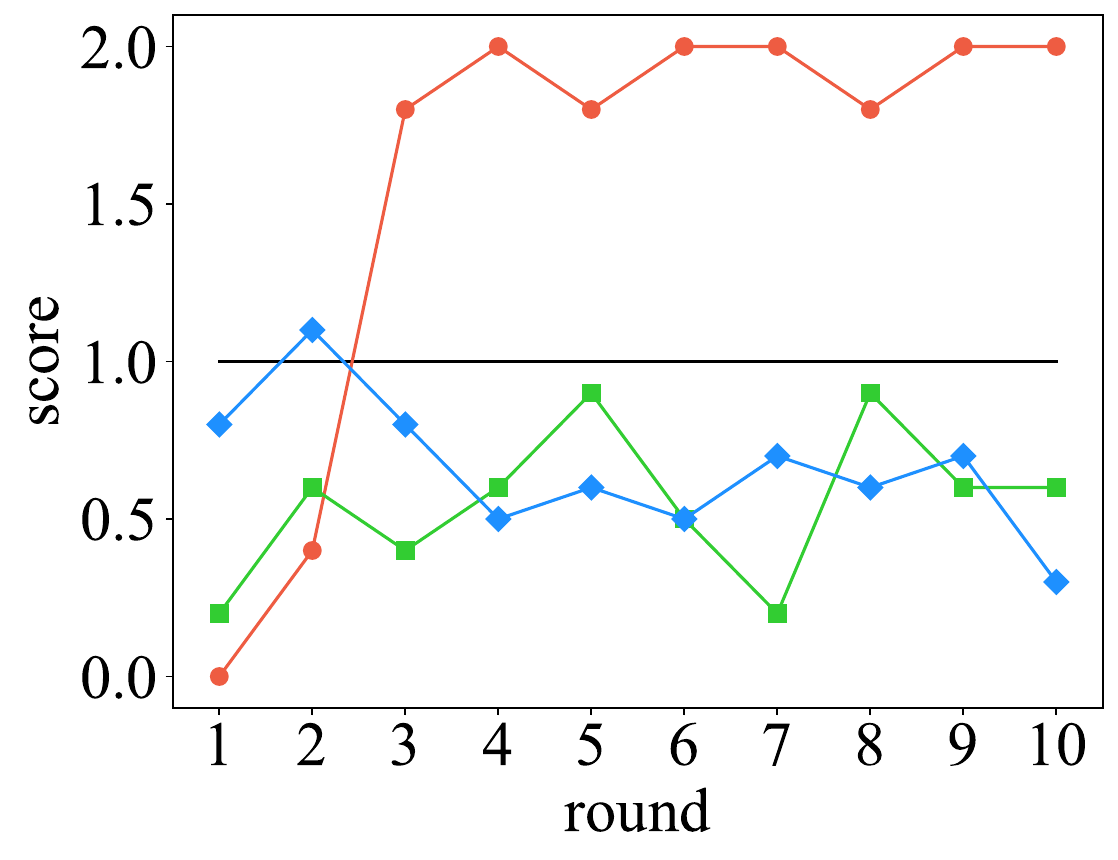} 
    \label{f4-2-1c}
    }
    \subfigure[copy]{
    \includegraphics[width=0.323\columnwidth]{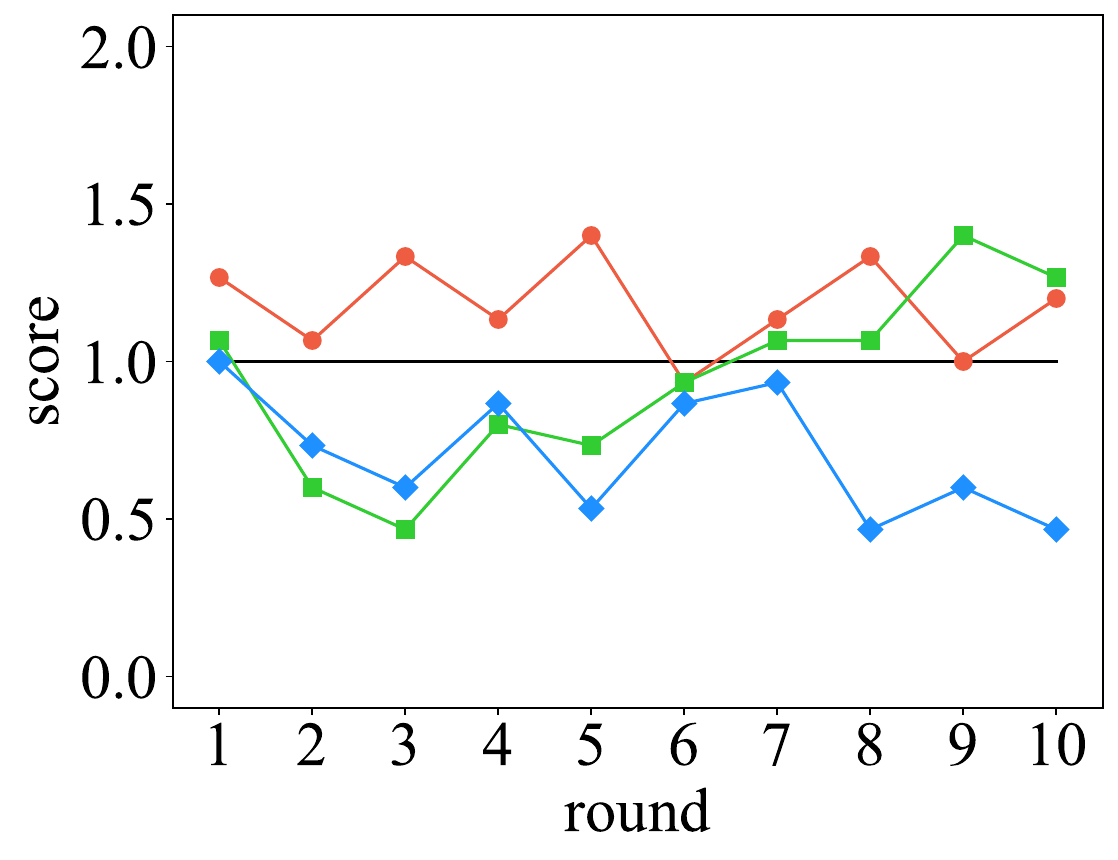}
    \label{f4-2-1d}
    }
    \subfigure[counter]{
    \includegraphics[width=0.323\columnwidth]{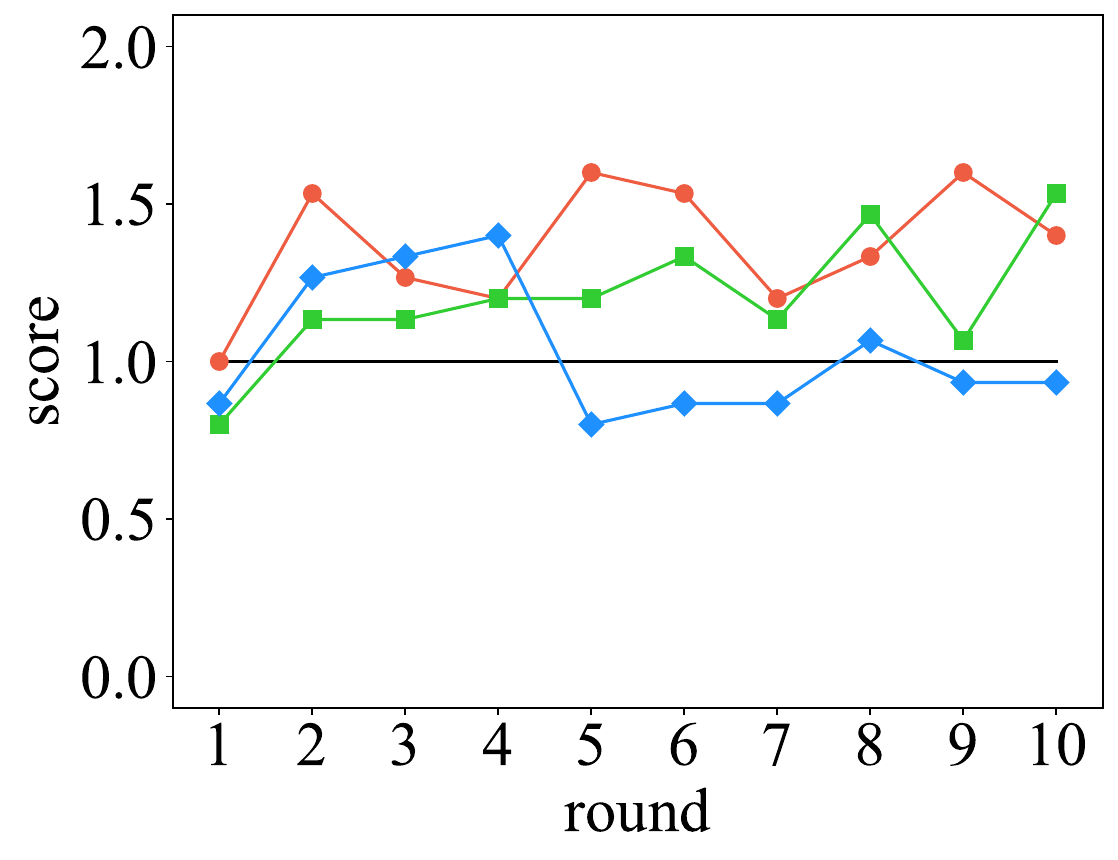} 
    \label{f4-2-1e}
    }
    \subfigure[sample]{
    \includegraphics[width=0.323\columnwidth]{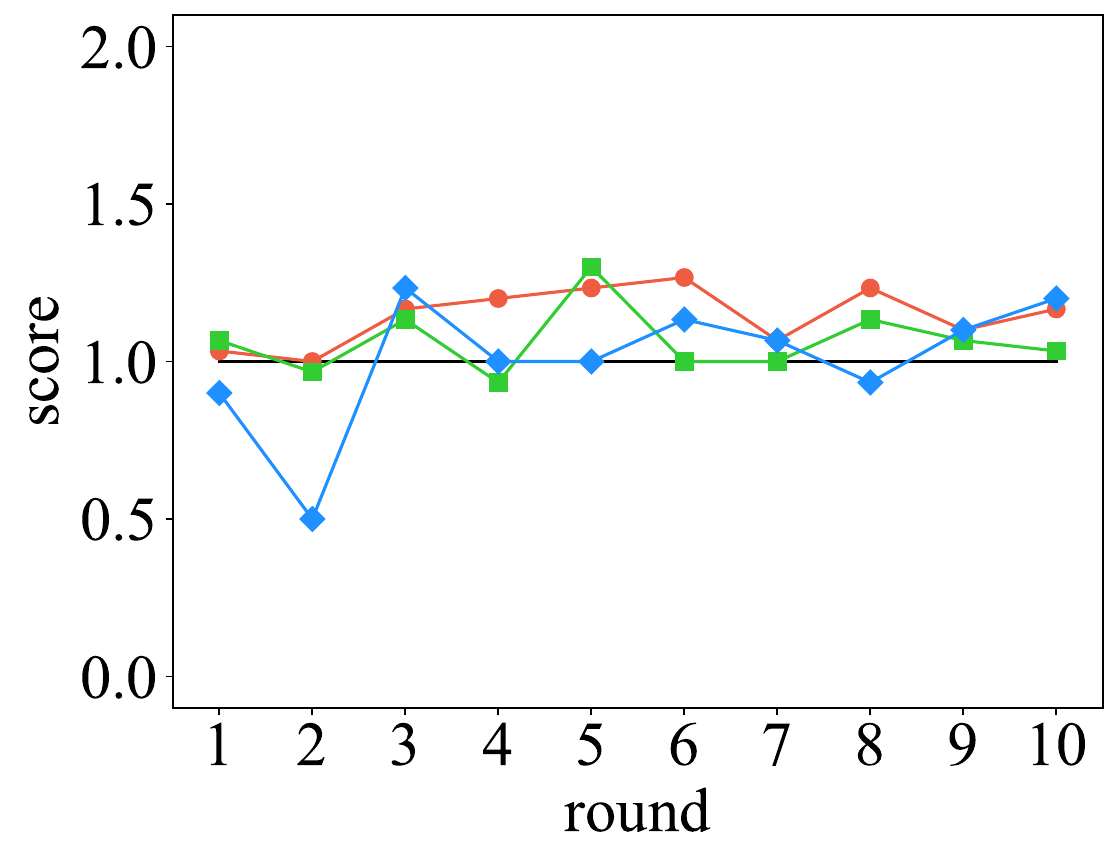} 
    \label{f4-2-1f}
    }
    \caption{Average payoff of LLMs for each round in R-S-P. }
    \label{f4-2-1}
\end{figure*}

\subsubsection{Analysis}

The experimental results are displayed in Table \ref{tab4-1-1}. When assigned common preferences (EQ and SI), all three LLMs made preference-consistent choices in all experiments, demonstrating the basic ability of LLMs to build clear desires from textual prompts. However, LLMs performed poorly when given uncommon preferences (CI and AL). Specifically, for the preference of CI, both GPT-3.5 and GPT-4 had sporadic errors, and the accuracy of GPT-3 was less than half; for the preference of AL, GPT-3.5 also made a large number of errors, while GPT-3 almost completely misunderstood AL (making the reference-consistent choice only once). The experimental results reveal significant differences in the ability of LLMs to build desires when assigned common / uncommon preferences. 

To further analyze the ability of LLMs to build a desire, we conducted a case study on the preference AL as illustrated in Fig. \ref{f4-1-1}. GPT-3's error stemmed from a lack of mathematical ability (confusion of numbers), which never occurred when GPT-3 is assigned a common preference. This seems to imply that the mathematical ability of LLMs assigned different preferences would be significantly different. GPT-3.5 incorrectly assumed that a higher joint income implied the maximization of the recipient's income (confusion of preferences), which can be attributed to the deviation of the built desire of GPT-3.5. GPT-4 performed well in this case, both analysis and choice were consistent with humans. 

\vspace{0.2cm}
\noindent\fbox{
  \parbox{230pt}{
    \textbf{Insight}: \emph{LLMs have the basic ability to build clear desires based on textual prompts, but struggle to build desires from uncommon preferences. \\We consider that providing more explicit and specific explanations of preferences may be helpful to LLMs when game experiments involve uncommon preferences. }
}}



\subsection{Can LLMs Refine Belief?}
\label{s4-2}

In game theory, a rational player needs to refine belief $\Omega_\mathcal{I}$ about uncertainty (e.g., opponent's action) from the game information $\mathcal{I}$. Essentially, refining belief is a process of synthesizing surface-level information into deeper insights. Because of the emphasis on decision-making in high uncertainty \cite{wellman2017mitigating}, game experiments in politics often examine players' ability to refine belief. 

Unfortunately, even for humans, refining belief can be a challenge. Therefore, it is meaningful to determine which types of beliefs LLMs can or cannot refine. 


\subsubsection{Game: Rock-Paper-Scissors}

Rock-Paper-Scissors (R-P-S) is a simultaneous, zero-sum game for two players. The rules of R-P-S are simple: rock beats scissors, scissors beat paper, paper beats rock; and if both players take the same action, the game is a tie. 

R-P-S is an ideal game to analyze LLMs' ability to refine belief. On the one hand, analyzing statistical patterns of non-random opponents' historical records can bring significant advantages in R-P-S \cite{Fisher2008RockPS}. On the other hand, for LLMs, R-P-S's preference (to win) is clear and the rules are simple: given the opponent's action, LLMs can always take the correct action based on the rules. Therefore, we consider that LLMs' performance in R-P-S can reflect LLMs' ability to refine belief. 

Specifically, in round $i$, the player's (my) action is noted as $a_m^i$ and the opponent's action is noted as $a_o^i$. After playing $t-1$ consecutive rounds with the same opponent, the historical records $\{ a_o^{< t}, a_m^{< t} \}$ can be considered as the game information $\mathcal{I}$ for refining belief $\Omega_\mathcal{I}$ in round $t$. So, the optimal strategy in round $t$ can be expressed as: 
\begin{equation}
\begin{split}
    \pi^*(a_m^t | \mathcal{I}) &= \pi^* (a^t | a_o^{< t}, a_m^{< t}) \\
    &= \operatorname*{argmax}_{a_m^t \in \mathcal{A}} \mathbb{E}_{a_o^{t} \sim p ( \Omega_{ \{ s_o^{< t}, a_m^{< t} \}} )}[D(a_o^t, a_m^{t})]. 
\end{split}
\label{e4-2-1}
\end{equation}

\noindent Since LLMs can grasp the preferences and rules of R-P-S, the difficulty of Eq. \ref{e4-2-1} lies in refining belief, i.e., $a_o^{t} \sim p ( \Omega_{ \{ s_o^{< t}, a_m^{< t} \}} )$. 


\begin{table}[t]
    \centering
    \begin{threeparttable}
    \resizebox{\columnwidth}{!}{
    \begin{tabular}{lcc}
    \toprule
    \bf Strategy & \bf Name & \bf Description \cr
    \midrule
        \multirow{1}*{$a_o^{t} = C$} & constant & remain constant \cr
        \midrule
        \multirow{2}*{$a_o^{t} = f(a_o^{< t})$} & loop-2 & loop between two actions \cr
         & loop-3 & loop among three actions \cr
        \midrule
        \multirow{2}*{$a_o^{t} = f(a_m^{< t})$} & copy & copy opponent's previous action \cr
         & counter & counter opponent's previous action \cr
        \midrule
        \multirow{1}*{$a_o^{t} \sim p(\mathcal{P}) $} & sample & sample in preference probability \cr
    \bottomrule
    \end{tabular}
    }
    \end{threeparttable}
    \caption{Summary of the opponent's strategy in R-S-P. }
    \label{tab4-2-1}
\end{table}

\subsubsection{Setup}

In international R-P-S programming competitions \cite{billings2000first}, a non-random opponent's action (in round $t$) is determined by the historical records $\{ a_o^{< t}, a_m^{< t} \}$ and the opponent's preference $\mathcal{P}$ as: 
\begin{equation}
    a_o^{t} \sim p(\mathcal{A} | a_o^{< t}, a_m^{< t}, \mathcal{P}). 
    \label{e4-2-2}
\end{equation}

\noindent Essentially, refining belief refers to making $p(\Omega)$ approach $p(\mathcal{A} | a_o^{< t}, a_m^{< t}, \mathcal{P})$. For a fine-grained analysis of the ability of LLMs to refine belief, we set up 4 simple opponent's patterns based on Eq. \ref{e4-2-2}, as shown in Table \ref{tab4-2-1}. $a_o^{t} = C$ is the basic pattern, evaluating the most basic refinement ability of LLMs. In this pattern, we conduct three experiments with the opponent's actions remaining constant as R, S, and P, respectively. $a_o^{t} = f(a_o^{< t})$ is determined by $a_o^{< t}$. Under the Markov assumption \cite{Puterman1994MarkovDP}, this pattern behaves as a loop. We conduct three loop-2 experiments (R-P, P-S, S-R) and one loop-3 experiment (R-P-S) in this pattern. $a_o^{t} = f(a_m^{< t})$ is determined by $a_m^{< t}$. Under the Markov assumption, we conduct two experiments in this pattern: copy / counter the player’s previous action $a_m^{t-1}$. $a_o^{t} \sim p(\mathcal{P}) $ is determined by the preference $\mathcal{P}$. To implement this pattern, we set a preference probability distribution of $(0.70, 0.15, 0.15)$ and conduct three experiments where the opponent has a preference for R, S, and P respectively, to take action by sampling in the distribution probability. 

To quantify the results of R-P-S, we set the payoff for a win as 2, for a tie as 1, and for a loss as 0. In each experiment, LLMs need to play 10 consecutive rounds of R-P-S against an opponent with a specific pattern, and the historical records are updated in time. Each experiment is repeated 10 times, and the temperature of LLMs is set to 0.7. 

\begin{figure}[t]
    \centering
    \subfigure[Analysis of GPT-3.5]{
    \includegraphics[width=234pt]{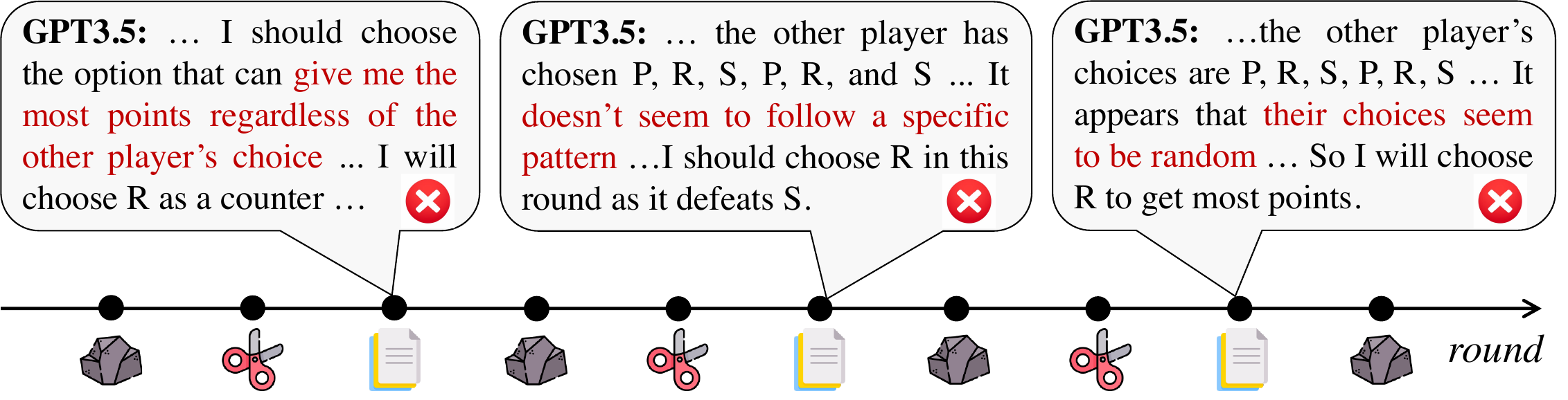}
    \label{f4-2-2a}
    }
    \subfigure[Analysis of GPT-4]{
    \includegraphics[width=234pt]{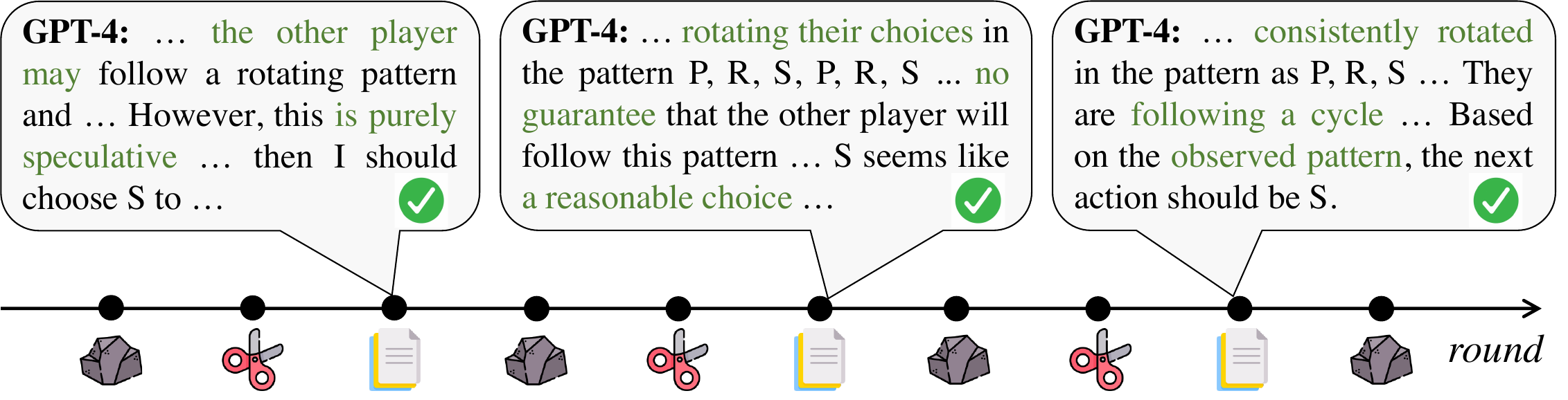} 
    \label{f4-2-2b}
    }
    \caption{Analysis of LLMs on loop-3. The symbols under the round axis indicate the opponent's action for each round. }
    \label{f4-2-2}
\end{figure}

\subsubsection{Analysis}

The average payoffs of each LLM are shown in Fig. \ref{f4-2-1}. Specifically, in the basic pattern (constant), GPT-3 performed close to random guessing, suggesting that GPT-3 lacked the basic ability to refine belief. In contrast, GPT-3.5's average payoff was significantly higher than random guessing and continued to rise; GPT-4 consistently took correct actions after approximately 3 rounds. In $a_o^{t} = f(a_o^{< t})$ pattern (loop-2, loop-3), GPT-3 and GPT-3.5 appeared to be capable of capturing some cyclical features, but they were unable to take correct actions. However, the performance of GPT-4 was exciting, with the update of historical records, the payoff was clearly rising. This led us to believe that GPT-4 can refine belief from this pattern. In $a_o^{t} = f(a_m^{< t})$ pattern (copy, counter), the situation was not ideal, GPT-4 seemed to have a slight advantage, but the overall performance of LLMs was not good enough. In $a_o^{t} \sim p(\mathcal{P})$ pattern (sample), the performance of all LLMs was similar to random guessing. Overall, LLMs are unable to refine belief well in most patterns, whereas for humans, the patterns involved in our experiments are quite easy to refine. 

For a more detailed analysis, we compared the analysis of GPT-3.5 and GPT-4 on loop-3, as shown in Fig. \ref{f4-2-2}. The analysis of GPT-3.5 demonstrated a lack of ability to refine belief. Even though GPT-3.5 expressed that the opponent's actions were P-R-S loops, it still believed that the opponent did not follow a specific pattern. The analysis of GPT-4, in contrast, was amazing, not only can GPT-4 summarize the opponent's pattern, but the tone gradually changed from uncertain to confident as the historical records were updated. 

\vspace{0.2cm}
\noindent\fbox{
  \parbox{230pt}{
    \textbf{Insight}: \emph{Currently, the ability of LLMs to refine belief is still immature and cannot refine belief from many specific patterns (even if simple). \\Therefore, we strongly recommend the cautious introduction of LLMs in game experiments that require refining complex belief. Nevertheless, the performance of GPT-4 in $a_o^{t} = f(a_o^{< t})$ pattern makes us look forward to more powerful LLMs in the future. }
}}

\subsection{Can LLMs Take Optimal Actions?}
\label{s4-3}

Taking optimal actions is the ultimate goal of a rational player in game theory, which requires the player to reason with known information (desire $D(\cdot)$ and belief $\Omega_\mathcal{I}$). Economics' obsession with optimal actions naturally makes game experiments in economics focus on analyzing players' actions \cite{kirzner1962rational,o2003economics}. 

However, for LLMs, there are various forms of combining desire and belief to take optimal actions, and it is unclear which form LLMs are more suitable in the game process. Here, we mainly explore the effect of the form of belief on LLMs taking optimal actions. 



\begin{figure}[t]
    \centering
    \subfigure[Payoff bimatrix]{
    \includegraphics[width=0.32\columnwidth]{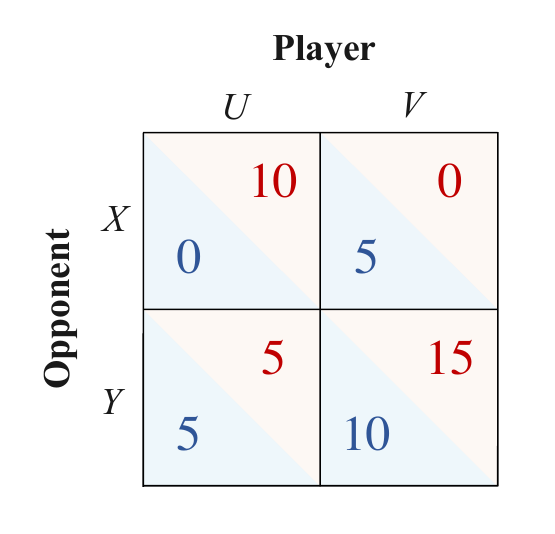}
    \label{f4-3-1a}
    }
    \centering
    \subfigure[Ideal game process]{
    \includegraphics[width=0.62\columnwidth]{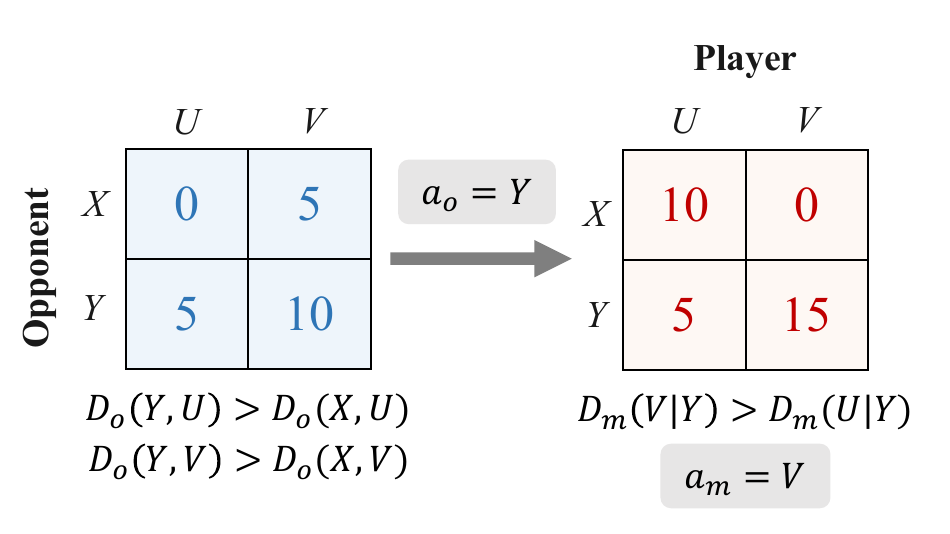}
    \label{f4-3-1b}
    }
    \caption{Overview of ring-network game, where \textcolor{darkred}{red numbers} / \textcolor{darkblue}{blue numbers} represent the player's and opponent's payoffs, and $D_m(\cdot)$ and $D_o(\cdot)$ represent the player's and opponent's desire functions. }
    \label{f4-3-1}
\end{figure}

\begin{table*}[t]
    \centering
    \begin{threeparttable}
    \begin{tabularx}{\linewidth}{p{0.4cm} >{\centering\arraybackslash}p{1.20cm} >{\centering\arraybackslash}p{1.20cm} >{\centering\arraybackslash}p{1.20cm}| >{\centering\arraybackslash}p{0.75cm} >{\centering\arraybackslash}p{0.75cm} >{\centering\arraybackslash}p{0.75cm} >{\centering\arraybackslash}p{0.75cm} >{\centering\arraybackslash}p{0.75cm} >{\centering\arraybackslash}p{0.75cm}| >{\centering\arraybackslash}p{1.20cm} >{\centering\arraybackslash}p{1.20cm} >{\centering\arraybackslash}p{1.20cm}}
    \toprule
    \multirowcell{3}{\\ \bf \#} & \multicolumn{3}{c|}{\bf Implicit Belief $\to$ Take Action} & \multicolumn{6}{c|}{\bf Explicit Belief $\to$ Take Action} & \multicolumn{3}{c}{\bf Given Belief $\to$ Take Action} \cr
    \cmidrule(lr){2-4} \cmidrule(lr){5-10} \cmidrule(lr){11-13}
     & GPT-3 & GPT-3.5 & GPT-4 & \multicolumn{2}{c}{GPT-3} & \multicolumn{2}{c}{GPT-3.5} & \multicolumn{2}{c|}{GPT-4} & GPT-3 & GPT-3.5 & GPT-4 \cr
    \cmidrule(lr){2-2} \cmidrule(lr){3-3} \cmidrule(lr){4-4} \cmidrule(lr){5-6} \cmidrule(lr){7-8} \cmidrule(lr){9-10} \cmidrule(lr){11-11} \cmidrule(lr){12-12}  \cmidrule(lr){13-13} 
    & $a_m$ & $a_m$ & $a_m$ & $a_o$ & $a_m$ & $a_o$ & $a_m$ & $a_o$ & $a_m$ & $a_m$ & $a_m$ & $a_m$ \cr
    \midrule
    (a) & 0.20 & 0.50 & 0.10 & 0.65 & 0.15 & 0.95 & 0.60 & 1.00 & 0.75 & 0.75 & 0.85 & 1.00 \cr
    (b) & 0.40 & 0.40 & 0.00 & 0.60 & 0.30 & 1.00 & 0.65 & 1.00 & 0.60 & 0.40 & 0.95 & 1.00 \cr
    (c) & 0.10 & 0.10 & 0.00 & 0.75 & 0.00 & 0.95 & 0.25 & 0.95 & 0.65 & 0.15 & 0.90 & 1.00 \cr
    (d) & 0.05 & 0.10 & 0.00 & 0.30 & 0.00 & 0.95 & 0.35 & 1.00 & 0.75 & 0.10 & 0.80 & 1.00 \cr
    \bottomrule
    \end{tabularx}
    \end{threeparttable}
    \caption{Performance of LLMs in different settings in the ring-network game. $a_o$ represents the accuracy of refining belief (the opponent's action), and $a_m$ represents the accuracy of taking the optimal action. }
    \label{tab4-3-1}
\end{table*}

\subsubsection{Game: Ring-Network Game}

The ring-network game is a game experiment that evaluates the rationality of taking actions in economics \cite{kneeland2015identifying}. In this research, we simplify it to a kind of 2 $\times$ 2 game (two players with two discrete actions). This game involves two players, the opponent and the player, whose preferences are to maximize their own payoff. In the game process, the opponent and the player need to take an action $a_o \in \{X, Y\}$ and $a_m \in \{U, V\}$, respectively. The payoff bimatrix $M$ consists of the opponent's matrix $M_o$ and the player's matrix $M_m$, as shown in Fig. \ref{f4-3-1a}, which specifies the payoffs of both sides for each combination of actions. 

The characteristic of the ring-network game is that players' optimal action is determined sequentially by the other players' optimal actions \cite{kneeland2015identifying}. The ideal game process is shown in Fig. \ref{f4-3-1b}, for the opponent, the payoff of $Y$ is always higher than $X$ regardless of the player’s actions. Therefore, the opponent's optimal action is always $Y$. For the player, the opponent's optimal action can be analyzed according to the opponent's payoff matrix $M_o$, so the player should be able to refine belief $\Omega$: $a_o = Y$. Then, the player can take the optimal action ($a_m=V$) based on belief and the player's payoff matrix $M_m$. According to the above analysis, the game information $\mathcal{I}$ is the payoff bimatrix $M$, and the player's optimal strategy can be expressed as: 
\begin{equation}
    \pi^*(a_m | \mathcal{I}) = \operatorname*{argmax}_{a_m \in \{U, V\}} [p(a_o | M) \cdot D_m(a_m| a_o, M) ], 
    \label{e4-3-1}
\end{equation}


\noindent where refining belief corresponds to $p(a_o | M)$ and taking the optimal action corresponds to $D_m(a_m| a_o, M)$. What we focus on is what form of bridging these two parts is more suitable for LLMs to take optimal action. 

\subsubsection{Setup}


Specifically, we set up three forms of combining belief based on Eq. \ref{e4-3-1} to analyze the performance of LLMs taking optimal actions in the ring-network game as: 
\begin{itemize}
    \item Implicit Belief $\to$ Take Action: \emph{We prompt LLMs in the dialogue to take the optimal action based on the payoff bimatrix directly, i.e., $\textsc{LLM}(a_m | M)$. In this form, LLMs need to autonomously transform this process into Eq. \ref{e4-3-1}. }
    \item Explicit Belief $\to$ Take Action: \emph{First, we prompt LLMs in the dialogue to refine belief (analyze the opponent's action) based on the payoff bimatrix, i.e., $\textsc{LLM}(a_o | M)$. Then, we continue the dialogue by prompting LLMs to take the optimal action based on the payoff bimatrix and the refined belief, i.e., $\textsc{LLM}(a_m| a_o, M)$. In this form, Eq. \ref{e4-3-1} is explicitly decoupled into two parts. }
    \item Given Belief $\to$ Take Action: \emph{The opponent's optimal action is explicitly provided to LLMs in the dialogue, and we prompt LLMs to take the optimal action based on the opponent's optimal action and payoff bimatrix, i.e., $\textsc{LLM}(a_m| a_o, M)$. In this form, LLMs only need to implement the second part of Eq. \ref{e4-3-1}. }
\end{itemize}

\noindent By analyzing the performance of LLMs in these three forms, we expect to obtain some caveats to help LLMs take optimal actions in game theory. 


\begin{figure}[t]
    \centering
    \includegraphics[width=215pt]{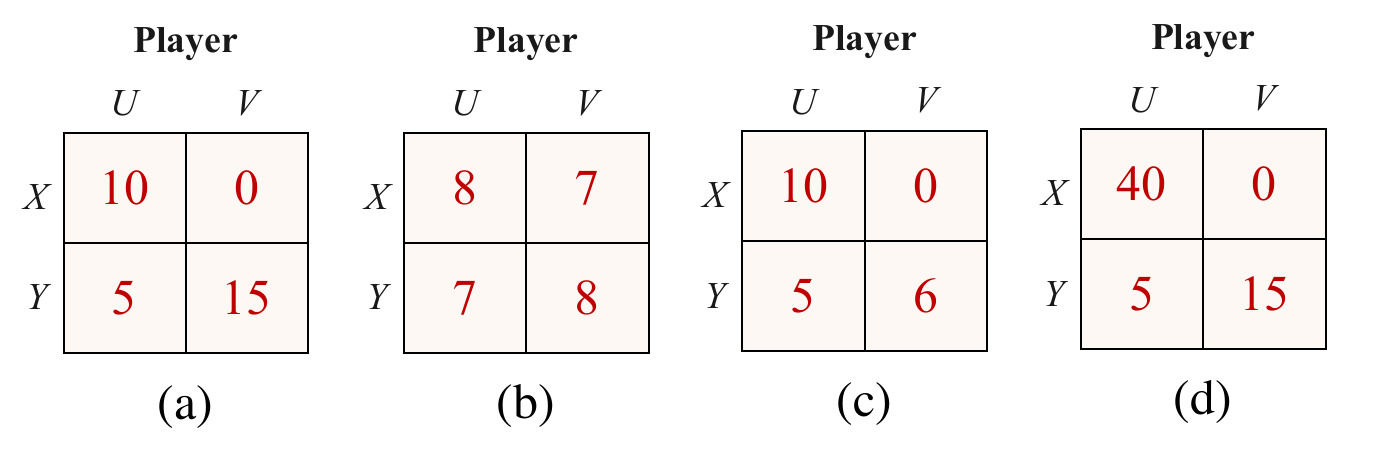}
    \caption{Setup of the player's payoff matrix. }
    \label{f4-3-2}
\end{figure}


In our experiments, in order to control the difficulty of refining belief, we keep the opponent's payoff matrix constant, which means the player's belief $\Omega$: $a_o = Y$ should remain constant. We set up different player's payoff matrices, as shown in Fig. \ref{f4-3-2}, to adjust the difficulty of taking the optimal action: (a) is the original setup; (b) reduces the difference in payoffs while keeping the expected payoffs to $a_m \in \{U, V\}$ constant; (c) increases the expected payoff for the incorrect action $a_m = U$; and (d) decreases the expected payoff for the correct action $a_m = V$. 


In practice, we find that LLMs are biased towards action names, e.g. GPT-3 prefers $U$ to $V$. In order to eliminate the influence of the bias of LLMs to take the optimal action, we swap the payoffs of $U$ and $V$ in the player's payoff matrix in Fig. \ref{f4-3-2}, to form a swapped payoff matrix, and we repeat the game 10 times each under the original and swapped payoff matrices and report the accuracy of the LLMs taking the optimal action. The temperature of LLMs is set to 0.7. 


\subsubsection{Analysis}

\begin{figure}[t]
    \centering
    \subfigure[Belief is overlooked: $p(a_m| a_o, M) \to p(a_m| M)$]{
    \includegraphics[width=234pt]{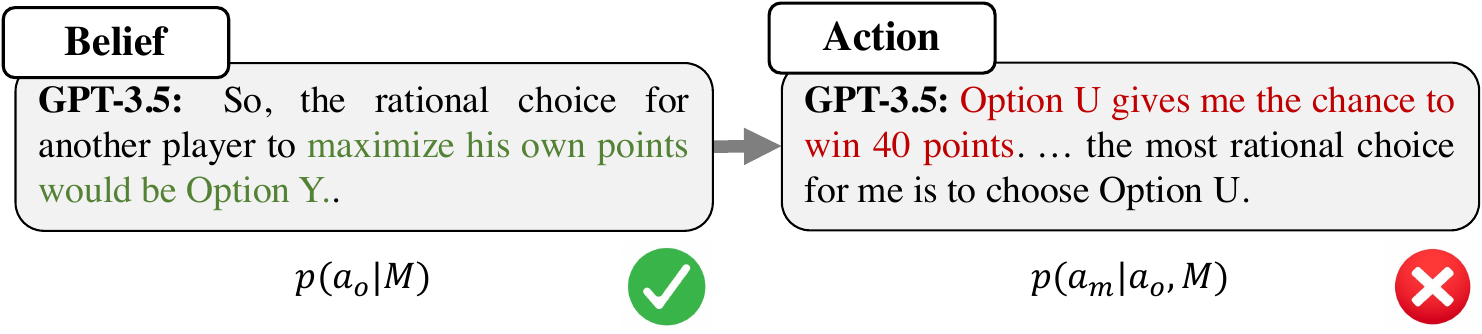}
    \label{f4-3-3a}
    }
    \subfigure[Belief is modified: $p(a_m| a_o, M) \to p(a_m| \hat{a}_o, M)$]{
    \includegraphics[width=234pt]{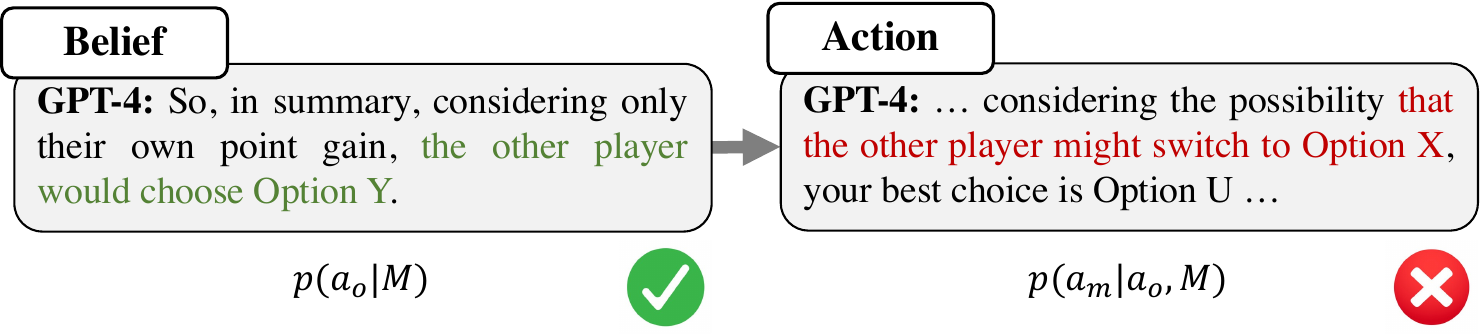} 
    \label{f4-3-3b}
    }
    \caption{Two cases of LLMs' inability to take optimal actions based on refined belief. }
    \label{f4-3-3}
\end{figure}

The performance of LLMs is shown in Table \ref{tab4-3-1}. Since GPT-3 performs poorly in all three forms, we mainly analyze the performance of GPT-3.5 and GPT-4. 

It is well known that human players' belief in game theory is implicit, so the form closest to humans taking optimal actions would be Implicit Belief $\to$ Take Action. However, all LLMs performed poorly in this form, and GPT-4 was almost completely unable to even take the optimal action. This reflected the capability gap between LLMs and humans, that was, LLMs cannot autonomously follow human behavior in the game process. In contrast, in the form of Explicit Belief $\to$ Take Action, by decomposing human behavior explicitly, the accuracy of LLMs to take the optimal action was significantly improved. This showed that LLMs were more suitable to take optimal actions in the explicit game process. This phenomenon was not unique to game theory, and many researches pointed out that explicitly decoupling human thoughts (think step-by-step) can significantly improve the performance of LLMs \cite{DBLP:conf/nips/Wei0SBIXCLZ22}. 


However, we were surprised that in the form of Explicit Belief $\to$ Take Action, LLMs were able to accurately refine belief (the accuracy for $a_o$ is above 0.95), but were unable to make the optimal action based on the refined belief well in subsequent dialogues, with the accuracy of GPT-4 being about 0.70 for $a_m$, and the accuracy of GPT-3.5 being even lower. As a comparison, we observed that when in the form of Given Belief $\to$ Take Action, GPT-4 was able to consistently take the optimal action, and GPT-3.5's accuracy also exceeded 0.80. Intuitively, LLMs are more suitable for taking optimal actions combining given belief rather than refined belief, even though the content of the two beliefs is the same. In order to explore the reasons, we conducted a detailed study on the error cases of GPT-3.5 and GPT-4 in the form of Explicit Belief $\to$ Take Action, and we summarized the two situations for LLMs' inability to take optimal actions based on refined belief as: 
\begin{itemize}
    \item Belief is overlooked: \emph{LLMs are confused by the game information and thus overlook the refined belief to take the optimal action in the subsequent dialogue. }
    \item Belief is modified: \emph{LLMs lack confidence in the refined belief and thus modify the refined belief to take the optimal action in the subsequent dialogue. }
\end{itemize}

\noindent The error cases are shown in Fig. \ref{f4-3-3}. In the first situation, as shown in Fig. \ref{f4-3-3a}, LLMs were confused by the expected payoff ($D_m(U) > D_m(V)$), and thus incorrectly equated $p(a_m| a_o, M)$ with $p(a_m| M)$. This occurred mainly on GPT-3.5. Observing the performance of GPT-3.5 in the form of Explicit Belief $\to$ Take Action, the accuracy of taking the optimal action was around 0.60 when the expected payoffs were the same (a and b), while the accuracy dropped to around 0.30 when the expected payoffs were different (c and d). In the second case, as shown in Fig. \ref{f4-3-3b}, LLMs modified the refined correct belief when taking the action due to lack of confidence, i.e., changing $p(a_m| a_o, M)$ to $p(a_m| \hat{a}_o, M)$. We found that modification of refined belief occurred more frequently on GPT-4. 



\vspace{0.2cm}
\noindent\fbox{
  \parbox{230pt}{
    \textbf{Insight}: \emph{We consider that LLMs do not have the ability to autonomously follow human behavior in the game process (in Fig. \ref{f1-1}). As a result, it is necessary to explicitly decouple human behavior for LLMs in game theory. \\However, even in the explicit game process, LLMs still appear to overlook / modify the refined belief. One possible solution is to transform the refined belief into the given belief in the dialogue. }
}}


\section{Conclusion}

The rapid development of LLMs leads us to believe that LLMs will eventually be integrated in all aspects of the human world, and therefore it is urgent to systematically analyze the capability boundaries of LLMs in various domains. In this research, we endeavor to systematically analyze LLMs in an important field of social science --- game theory. Our experiments evaluate to what extent LLMs can serve as rational players from three aspects and find some weaknesses of LLMs in game theory. 

As an early attempt to analyze LLMs in the context of game theory, our research has some limitations. For example, the difficulty of the game we selected is relatively low, not close enough to the real game scenarios; our perspective of analyzing the ability of LLMs is not rich enough, only considering the principle of rationality; our process of analyzing the game experiments is relatively rough and lacks more comparative and ablative experiments; and so on. 

In the future, we hope to apply LLMs more deeply in game theory. For example, LLMs as multi-agents in the game, the confrontation between humans and LLMs in the game, and the dynamic game in real scenarios. When it is clearly recognized that LLMs have specific ability deficiencies, it is also a promising research direction to design a targeted training process to improve the specific ability of LLMs. In conclusion, the research on LLMs in the context of game theory is still in a very preliminary stage, and a lot of exploratory researches are required. 

\section{Acknowledgments}

This work was supported by the Shanghai Municipal Science and Technology Major Project (2021SHZDZX0102), and the Fundamental Research Funds for the Central Universities. 

\bibliography{reference}

\appendix

\end{document}


\maketitle

\section{Prompts}

\subsection{Dictator Game}

The prompt is listed in Table \ref{tab1},including template for game instruction, four different preferences and corresponding incomes.

\begin{table*}[t]
    \centering
    
    \begin{tabular}{cp{13cm}}
    \toprule
         Template & \{\textit{Preference}\} You are playing a game with another player. Now you can determine the incomes for you and the other player. You can choose option 'X' or option 'Y'. If you choose 'X', you get \{\} and other player get \{\}. If you choose 'Y', you get \{\} and the other player gets \{\}. Q: Which option do you choose, 'X' or 'Y'? Give your analysis and output your final option. A:\cr
    \midrule
        AL preference & You have a stronger preference for another player’s interest and maximize another player’s income. \quad (\$500, \$100) \cr
    \midrule
        CI preference & You have a stronger preference for common interest and maximize the joint income.\quad (\$400, \$300)\cr
    \midrule
        SI preference & You have a stronger preference for your own interest and maximize your own income.\quad (\$100, \$500)\cr
    \midrule
        EQ preference & You have a stronger preference for fairness between players and hate inequality.\quad (\$300, \$300)\cr
    \bottomrule
    \end{tabular}
    
    \caption{Prompt for Dictator game }
    \label{tab1}
\end{table*}

\subsection{Rock-Paper-Scissors Game}

The prompt is listed in Table \ref{tab2}. There are four parts:
\begin{itemize}
    \item rule description: explain how to calculate points according to two players' action
    \item Analyze template: prompt LLMs to generate analysis of the other player's action pattern
    \item Choose template: instruct LLMs to choose the action
    \item Feedback template: provide the feedback after each round
\end{itemize}

\begin{table*}[t]
    \centering
    
    \begin{tabular}{cp{14cm}}
    \toprule
         Rule & You are playing a game with another player. In this game, You can choose between three options: 'R', 'S', 'P'. Your only target is to get as many points as possible. The rules of the game are as follows: If you choose 'R' and the other player chooses 'R', then you win 1 points and the other player wins 1 points. If you choose 'R' and the other player chooses 'S', then you win 2 points and the other player wins 0 points. If you choose 'R' and the other player chooses 'P', then you win 0 points and the other player wins 2 points. If you choose 'S' and the other player chooses 'S', then you win 1 points and the other player wins 1 points. If you choose 'S' and the other player chooses 'P', then you win 2 points and the other player wins 0 points. If you choose 'S' and the other player chooses 'R', then you win 0 points and the other player wins 2 points. If you choose 'P' and the other player chooses 'P', then you win 1 points and the other player wins 1 points. If you choose 'P' and the other player chooses 'R', then you win 2 points and the other player wins 0 points. If you choose 'P' and the other player chooses 'S', then you win 0 points and the other player wins 2 points.\cr
    \midrule
        Analyze & Your are playing round \{\}, Please give your analysis based on another player's selection pattern in previous rounds. \cr
    \midrule
        Choose & Q: Your are playing round \{\}, which option do you choose, 'R', 'S' or 'P'? Output your final option, e.g. I choose 'option'. A:\cr
    \midrule
        Feedback & In round {}, you chose '\{\}' and the other player chose '\{\}'. Thus, you won \{\} points and the other player won \{\} points.\cr
    \bottomrule
    \end{tabular}
    
    \caption{Prompt for Rock-Paper-Scissors game }
    \label{tab2}
\end{table*}

\subsection{Ring-Network Game}

The prompt is listed in Table \ref{tab3}. There are five parts:
\begin{itemize}
    \item rule description: explain the payoff matrices of two players
    \item Analyze 1 template: prompt LLMs to generate analysis of the other player's action
    \item Choose 1 template: instruct LLMs to choose the other player's action
    \item Analyze 2 template: prompt LLMs to generate analysis of their own action
    \item Choose 2 template: instruct LLMs to choose the their own action
\end{itemize}

\begin{table*}[t]
    \centering
    
    \begin{tabular}{cp{14cm}}
    \toprule
       Rule  &  You are playing a game with another player. In this game, you can choose between Option U and Option V. The other player can choose between Option X and Option Y. You and the other player simultaneously choose the option. Your payoffs are as follows: If you choose Option U and the other player chooses Option X, then you win \{\} points. If you choose Option U and the other player chooses Option Y, then you win \{\} points. If you choose Option V and the other player chooses Option X, then you win \{\} points. If you choose Option V and the other player chooses Option Y, then you win \{\} points. The other player's payoffs are as follows: If you choose Option U and the other player chooses Option X, then the other player wins \{\} points. If you choose Option U and the other player chooses Option Y, then the other player wins \{\} points. If you choose Option V and the other player chooses Option X, then the other player wins \{\} points. If you choose Option V and the other player chooses Option Y, then the other player wins \{\} points.\cr
    \midrule
        Analyze 1 & You know the other player is rational to choose one option and maximize his own points. Based on game rule, give your analysis of the other player's choice. \cr
    \midrule
        Choose 1 & Q: Which option do you think the other player will choose, Option X or Option Y? Only Output final option. A:\cr
    \midrule
        Analyze 2 & Based on analysis of the other player's choice, analyze your choice to maximize your own points.\cr
    \midrule
        Choose 2 & Q: Which option do you choose, Option U or Option V? Only Output your final option. A:\cr
    \bottomrule
    \end{tabular}
    
    \caption{Prompt for Ring-Network game }
    \label{tab3}
\end{table*}

\section{Generation Examples}

\subsection{Dictator Game}

The correct generated examples of three models are shown in Table \ref{tab4} and  the wrong examples are shown in Table \ref{tab5}. 

\begin{table*}[t]
    \centering
    
    \begin{tabular}{cp{14cm}}
    \toprule
       GPT-3  &  I would choose option 'X' as it provides 500 dollars in income, compared to only 400 dollars in income under option 'Y'. This will provide me with the highest possible return.\cr
    \midrule
        GPT-3.5 &  As an AI, my main objective is to maximize my own income. Based on the given options, it is clear that option 'X' would result in a higher income for me (\$500) compared to option 'Y' (\$400). Therefore, I would choose option 'X' to maximize my own income.\cr
    \midrule
        GPT-4 & Given that I am only focused on my own interests and maximizing my own income, I would choose option 'X'. This is because option 'X' provides me with 500 dollars, which is 100 dollars more than what I would get with option 'Y'. The other player's income is not a concern to me in this scenario. Hence, my final choice would be option 'X'.\cr
    \bottomrule
    \end{tabular}
    \caption{Correct examples of three models in SI-CI setting.}
    \label{tab4}
\end{table*}

\begin{table*}[t]
    \centering
    
    \begin{tabular}{cp{14cm}}
    \toprule
       GPT-3  &  Option Y should be chosen as it is the option that maximizes the joint income. By choosing option Y, the two players would have a combined total of 600 dollars, whereas option X only results in a combined total of 700 dollars. Therefore, option Y should be chosen as it is the optimal option that maximizes the joint income.\cr
    \midrule
        GPT-3.5 & If I choose option 'X', I would receive 400 dollars and the other player would receive 300 dollars. This option would result in a total income of 700 dollars. If I choose option 'Y', I would receive 100 dollars and the other player would receive 500 dollars. This option would result in a total income of 600 dollars. Analyzing the options, it is clear that option 'Y' leads to a higher total income of 600 dollars. Therefore, my final option would be 'Y'. \cr
    \midrule
        GPT-4 & I choose option 'Y'. If we sum up the incomes, option 'X' gives a total of 700 dollars (400+300), while option 'Y' gives a total of 600 dollars (100+500). However, my aim here is to maximize the joint income. Therefore, we get more total income in option 'X'. But I also care about common interest and fair distribution of income. If I choose option 'Y', the other player gets a higher income, which is fairer and can be seen as common interest. Considering both aspects, I will choose option 'Y'.\cr
    \bottomrule
    \end{tabular}
    \caption{Wrong examples of three models in AL-CI setting.}
    \label{tab5}
\end{table*}

\subsection{Rock-Paper-Scissors Game}

The several generated examples of three models are shown in Table \ref{tab6}.

\begin{table*}[t]
    \centering
    
    \begin{tabular}{cp{14cm}}
    \toprule
       GPT-3  &   Based on the other player's selection pattern in previous rounds, it would be prudent to choose 'P' as it stands the best chance of earning 2 points while the other player has the least chance of earning 2 points.\cr
    \midrule
        GPT-3.5 &  Based on the previous rounds, the other player has consistently chosen 'S' in each round. Given this pattern, it is highly likely that they will continue to choose 'S' in round 10. To maximize my points, I need to choose an option that will give me the highest score against their expected choice of 'S'. Since 'R' beats 'S' and gives me 2 points, I will choose 'R' once again in round 10. This way, if the other player does choose 'S', I will win 2 points and they will win 0 points.\cr
    \midrule
        GPT-4 & Throughout the game, the other player has consistently chosen 'S'. My choice of 'R' has allowed me to consistently win 2 points. Even though they might be inclined to change their strategy in this final round, their pattern suggests they might stick with 'S'. If they opt for 'R', choosing 'P' would be the best option for me. However, given their consistent pattern, it seems best to stick to 'R'. I choose 'R'.\cr
    \bottomrule
    \end{tabular}
    \caption{Several examples of three models.}
    \label{tab6}
\end{table*}

\subsection{Ring-Network Game}

The several generated examples of three models are shown in Table \ref{tab7}.

\begin{table*}[t]
    \centering
    
    \begin{tabular}{cp{14cm}}
    \toprule
       GPT-3  &  The other player will choose Option Y since it gives them the highest payoff of 10 points. This means that the best option for you is to choose Option U, since it gives you the highest payoff of 10 points. \newline \textit{Option Y.} \newline My choice is Option U, since it gives me the highest payoff of 10 points. \newline \textit{Option U.}\cr
    \midrule
        GPT-3.5 &  Based on the payoffs, we can analyze the other player's choice based on their rationality and the goal of maximizing their own points. If I choose Option U, the other player's best response would be to choose Option Y since it gives them 5 points, compared to 0 points if they choose Option X.  Similarly, if I choose Option V, the other player's best response would be to choose Option X since it gives them 5 points, compared to 10 points if they choose Option Y. Therefore, the other player's rational choice would be to choose Option Y if I choose Option U, and Option X if I choose Option V. They will make their decision based on the option that gives them more points. To summarize, the other player's choice depends on my choice. If I choose Option U, they will choose Option Y. If I choose Option V, they will choose Option X. \newline \textit{The other player will choose Option Y.} \newline Based on the analysis of the other player's choice, if they choose Option Y, I should choose Option U to win 5 points. If they choose Option X, I should choose Option V to win 0 points. To maximize my own points, I will choose Option U if the other player chooses Option Y, and I will choose Option V if the other player chooses Option X. This way, I can ensure that I am making the choice that gives me the maximum possible points in each scenario. \newline \textit{I choose Option U.}\cr
    \midrule
        GPT-4 & Given that the other player is rational and seeks to maximize their own points, they will likely choose Option Y. This is because their maximum possible gain from Option X is only 5 points (when you choose option V), but they could gain 10 points from Option Y if you choose option V.  However, it's also important to note that they may anticipate your choices. If they think you'll choose Option U (which gives you more points on average), they might decide to select Option Y to get 5 points since Option X will give them 0 points.  So in summary, considering only their own point gain, the other player would choose Option Y. But if they are considering your choices as well, they might change their strategy. \newline \textit{Option Y} \newline Given that the other player is likely to choose Option Y, you should choose Option U to maximize your points. When the other player chooses Option Y, your payoff with Option U is 5 points, and with Option V it is 15 points. However, considering the possibility that the other player might switch to Option X, your best choice is Option U, because if they choose Option X and you choose Option U, you get 10 points, which is higher than the 0 points you would get if you chose Option V. So, choosing Option U gives you a more balanced potential payoff. \newline \textit{Option U}\cr
    \bottomrule
    \end{tabular}
    \caption{Several examples of three models.}
    \label{tab7}
\end{table*}